%% file: root.tex
\documentclass[letterpaper, 10 pt, conference]{ieeeconf}  

\IEEEoverridecommandlockouts                              

\usepackage[resetlabels]{multibib}
\usepackage{subfiles}
\usepackage{csquotes}
\usepackage{gensymb}
\usepackage{pifont}
\usepackage{textcomp}
\usepackage[table]{xcolor}
\usepackage{todonotes}
\usepackage{blindtext}
\usepackage{soul}
\usepackage{graphicx}
\usepackage{mathtools}
\usepackage{multirow}
\usepackage{flushend}
\usepackage{booktabs}
\usepackage{array}
\usepackage{tabularx}
\usepackage{threeparttable}
\usepackage{caption}
\usepackage{float}
\usepackage{amsmath} 
\usepackage{amssymb} 
\usepackage{amsfonts}
\usepackage{bm} 
\usepackage{siunitx}
\usepackage{cancel}
\usepackage{microtype}
\usepackage{cite}
\usepackage[export]{adjustbox}
\usepackage[hang,flushmargin]{footmisc}
\usepackage{cuted}
\newcites{S}{References}
\usepackage[breaklinks,colorlinks]{hyperref}
\usepackage{pgfplots}
\pgfplotsset{compat=1.17}
\usepackage[table]{xcolor}

\definecolor{Gray}{gray}{0.9}

\usepackage[capitalize]{cleveref}
\crefname{section}{Sec.}{Secs.}
\Crefname{section}{Section}{Sections}
\Crefname{table}{Table}{Tables}
\crefname{table}{Tab.}{Tabs.}

\definecolor{mygreen}{HTML}{00A64F}
\definecolor{myred}{HTML}{ED1B23}

\newcommand{\secref}[1]{Sec.~\ref{#1}}
\renewcommand{\eqref}[1]{Eq.~(\ref{#1})}
\newcommand{\figref}[1]{Fig.~\ref{#1}}
\newcommand{\tabref}[1]{Tab.~\ref{#1}}

\newcommand{\net}{StreamRPS}

\title{\LARGE \bf
Real-Time Radar Semantic and Panoptic Scene Understanding 
}
\author{Santosh Parit$^{1,*}$, Julia Hindel$^{1,*}$, Rohit Mohan$^{1}$, Georgi Pramatarov$^{2}$, Paul Newman$^{2}$, \\
Daniele De Martini$^{2,\dag}$ and Abhinav Valada$^{1,\dag}$
\vspace{0.5em}
\thanks{$^{1}$University of Freiburg, $^{2}$University of Oxford.}%
\thanks{$^*$ Equal contribution, $^\dag$ Equal supervision.}%
\thanks{This work was funded by the Deutsche Forschungsgemeinschaft (DFG, German Research Foundation) -- SFB 1597 -- 499552394 and J.H. acknowledges travel support from ELISE (GA no. 951847) and ELSA - European Lighthouse on Secure and Safe AI funded by the European Union under GA no. 101070617.}%
}

\begin{document}

\maketitle
\thispagestyle{empty}
\pagestyle{empty}

\begin{abstract}

Robust scene understanding under varying environmental conditions is essential for safe autonomous driving. While cameras and LiDAR degrade in adverse weather and poor lighting, Frequency-Modulated Continuous-Wave (FMCW) spinning radar provides dependable 360° perception at extended ranges with minimal compute cost. However, real-time FMCW radar scene understanding is challenging due to complex signal returns, limited temporal resolution, and scarce annotated datasets. We introduce \net, a novel framework for streaming FMCW radar panoptic segmentation, comprising an automated label-generation pipeline via LiDAR scene completion and a real-time streaming network for semantic and panoptic segmentation. 
Our encoder-decoder streaming architecture comprises a radar-specific
hierarchical Transformer encoder and task-specific decoders
for semantic or panoptic prediction. Our encoder introduces
a stem to capture range-dependent signatures, anisotropic patch embeddings to preserve angular resolution, and axial polar attention to separately model the periodic azimuth and bounded range axes. We further introduce azimuth roll rotation augmentation to simulate ego-vehicle rotation during training.
Comprehensive experiments on the SAX and BOREAS datasets validate both our auto-labeling methodology and streaming segmentation network against established baselines. We outperform the strongest baseline by at least 1.4pp while processing sectors at over 400 FPS in a 4-sector streaming setting. We additionally provide manual annotations for 100 SAX radar scans, enabling independent evaluation of the entire framework.
\end{abstract}


\input{sections/01_introduction}
\input{sections/02_related_works}
\input{sections/03_methodology}

\input{sections/04_experiments}
\input{sections/05_conclusion}


\footnotesize
\bibliographystyle{IEEEtran}
\bibliography{references}

\end{document}

%% file: sections/01_introduction.tex
\section{Introduction}\label{sec:introduction}
Precise environmental perception is essential for autonomous vehicles to navigate safely~\cite{luz2026latent}. This perception relies on sensor data from cameras, LiDAR, and radar, each offering distinct capabilities. However, cameras are sensitive to extreme lighting, and both cameras and LiDAR can suffer degraded performance in adverse weather, including rain, snow, and fog. In contrast, Frequency-Modulated Continuous-Wave (FMCW) radar offers greater resilience to these conditions, making it a promising modality for dense scene understanding. In particular, spinning FMCW radar provides $360^\circ$ angular coverage from a single sensor while maintaining low data-transfer bandwidth and memory requirements~\cite{prannay2020}.

Despite these sensing advantages, the complexity and noise characteristics of FMCW radar make manual annotation for training segmentation models prohibitively expensive~\cite{carrada20}. Existing approaches therefore rely on camera~\cite{carrada20,sheeny2021radiate} or LiDAR~\cite{li2025get} data for supervision. However, camera-based supervision is constrained by limited viewing angles, while sparse LiDAR measurements leave many regions of the radar grid unlabeled. To address these limitations, we leverage LiDAR scene completion to generate dense radar annotations. The resulting annotations cover $8$-$12$ semantic classes, exceeding the $3$-$4$ classes used in previous studies~\cite{carrada20,li2025get,fang2023,Rebut_2022_CVPR}.

\begin{figure}
    \centering
    \includegraphics[width=0.45\textwidth]{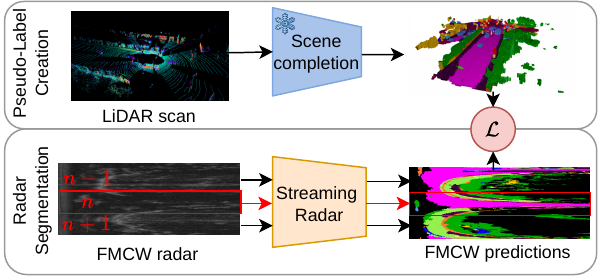}
    \caption{\net~yields dense semantic and panoptic radar labels using a pre-trained LiDAR scene completion network. The streaming FMCW radar segmentation architecture is trained on generated label sectors and operates in real time to deliver semantic or panoptic scene understanding.}
    \label{fig:teaser}
    \vspace{-0.6cm}
\end{figure}

Beyond dense supervision, effective radar segmentation requires architectures tailored to the sensor's distinct measurement characteristics. Existing approaches address these requirements through specialized convolutional and attention mechanisms~\cite{Dalbah_2024_WACV,Zhang_2023_CVPR,Ouaknine_2021_ICCV}. However, many of these methods rely on Doppler velocity information, which is often unavailable in spinning FMCW radar data. A further challenge stems from the reliance of existing spinning-radar segmentation approaches on complete $360^\circ$ scans. At relatively low scan rates ($\SI{4}\,$Hz), waiting for a complete sweep delays predictions. Streaming-capable architectures instead process incoming angular sectors throughout scan acquisition, enabling scene updates before the sweep is complete. Furthermore, to the best of our knowledge, prior work has not explored panoptic segmentation for spinning FMCW radar. Panoptic segmentation distinguishes individual objects within semantic categories through instance-level identifiers~\cite{mohan25}, providing object-level information for motion forecasting and navigation decisions. We therefore propose a specialized streaming architecture that supports both semantic and panoptic scene understanding.\looseness=-1

We present \net~(\textbf{Stream}ing \textbf{R}adar \textbf{P}anoptic \textbf{S}egmentation), a framework for generating dense radar annotations and enabling streaming semantic and panoptic scene understanding (\figref{fig:teaser}). Our labeling pipeline leverages the pre-trained LiDAR scene completion network PaSCo~\cite{cao2024pasco} to infer dense scene geometry and semantics from sparse LiDAR observations. We align these predictions with the radar's sequential acquisition through azimuth-based synchronization, then construct and refine instance annotations in polar radar coordinates. For streaming segmentation, we propose an encoder-decoder architecture comprising a radar-specific hierarchical Transformer encoder and task-specific decoders for semantic or panoptic prediction. Our encoder introduces a stem that combines positional information, directional filtering, and temporal feature weighting to capture range-dependent radar signatures. Anisotropic patch embeddings retain finer angular resolution, while axial polar attention distinguishes the periodic azimuth axis from the bounded range axis. We complement these encoder contributions with azimuth roll augmentation, which simulates ego-vehicle rotation through circular shifts during training. 

We evaluate both the label-generation pipeline and the radar segmentation network on the SAX~\cite{SAX20} and Boreas~\cite{burnett_ijrr23} datasets. For SAX, we additionally create ground-truth labels by manually annotating $350$ LiDAR scans and projecting these annotations onto $100$ radar scenes using sensor calibration. These annotations provide held-out ground truth independent of our automatically generated labels. \net{} achieves the highest semantic mIoU among the evaluated baselines across all tested sector configurations and the highest panoptic quality in the four-sector setting on both datasets. The network further maintains real-time inference in both full-scan and streaming settings.

In summary, our main contributions are:
\begin{itemize}
\item A novel auto-label generation pipeline for obtaining dense semantic and panoptic FMCW radar labels.
\item A streaming FMCW radar segmentation network supporting both semantic and panoptic predictions, demonstrating state-of-the-art performance on the SAX and Boreas benchmarks.
\item Manually curated annotations comprising $350$ LiDAR frames and $100$ calibrated radar scans from the SAX dataset.
\item Public release of code and trained models upon acceptance.
\end{itemize}

%% file: sections/02_related_works.tex
\section{Related Work}\label{sec:related-work}

{
\noindent\textit{Radar Segmentation Label Generation}:
To reduce manual annotation effort, existing methods generate radar labels using camera data, LiDAR data, or both. CARRADA~\cite{carrada20} associates mean-shift radar clusters with object categories using camera Mask R-CNN predictions and temporal tracking, yielding sparse annotations and dense object masks. RAMP-CNN~\cite{geo21ramp_cnn} uses camera-based object detection and depth estimation to generate radar training labels. Heidbrink et al.~\cite{heidbrink24} combine camera segmentation with LiDAR-based object localization for radar annotation. RADIal~\cite{Rebut_2022_CVPR} uses synchronized camera, LiDAR, and high-definition radar data to annotate vehicles and free space. For spinning FMCW radar, RSS-Net~\cite{prannay2020} projects camera predictions onto LiDAR points and transfers their labels to radar using calibration and motion compensation.

LiDAR-only supervision has also been explored for spinning FMCW radar. Li~et~al.~\cite{li2025get} use pre-trained 3D segmentation models, remove ground points, and group semantic labels into four categories. They refine predictions using DBSCAN clustering and SVD-based structural analysis before projection into radar coordinates. The structural analysis distinguishes building-like surfaces from vegetation to correct mislabeled points. Although these procedures improve label consistency, projection-based supervision can remain spatially incomplete. Our framework instead uses LiDAR scene completion to generate dense semantic and panoptic radar annotations across $8$-$12$ categories.}\looseness=-1

{\parskip=2pt
\noindent\textit{2D Radar Semantic Segmentation}: 
Spinning FMCW radars have been studied for localization and place recognition~\cite{venon22, lisus25}, while semantic segmentation has received comparatively limited attention. Existing approaches include the U-Net-based method of Li~et~al.~\cite{li2025get} and RSSNet~\cite{prannay2020},  which combines an ASPP-based decoder with class weighting and temporal scan stacking to improve segmentation of small and dynamic objects.
In contrast, extensive research has focused on non-spinning FMCW radars. Many approaches exploit complementary range-angle (RA) and range-Doppler (RD) representations. On CARRADA dataset~\cite{carrada20}, MVNet~\cite{Ouaknine_2021_ICCV} uses dual encoder-decoder networks with shared latent features, while TMVA-Net~\cite{Ouaknine_2021_ICCV} adds temporal aggregation through 3D convolutions. 
PeakConv~\cite{Zhang_2023_CVPR} introduces peak-inspired convolutional receptive fields, which AdaPKC~\cite{li2024adapkc} adapts to varying radar responses. 
Transformer-based approaches include TransRSS~\cite{zou2023}, TransRadar~\cite{Dalbah_2024_WACV} with adaptive directional attention and class-agnostic localization supervision, and TARSS-Net~\cite{zhang24tarssnet} with learned temporal relations. 
These methods collectively rely on Doppler information via RD representations. For the related task of panoptic segmentation, SemRaFiner~\cite{zeller2024semrafiner} operates on sparse radar point clouds and also uses Doppler information. Extracting Doppler from spinning FMCW radar, however, remains challenging~\cite{lisus25}, creating a gap between the rich literature on non-spinning radars and the underexplored semantic understanding of spinning systems. We bridge this gap by establishing that high-quality panoptic and semantic segmentation is possible without Doppler. Critically, this is the first work to tackle panoptic segmentation for FMCW radar.}

{\parskip=2pt
\noindent\textit{Streaming Segmentation for Spinning Sensors}:  
While streaming segmentation has not yet been explored for spinning FMCW radar, prior work in LiDAR perception has investigated this direction. Han et al.~\cite{han2020} introduce streaming object detection, while STROBE~\cite{frossard2021} processes LiDAR packets using a bird's-eye-view (BEV) convolutional backbone and multi-scale spatial memory. However, representing wedge-shaped sectors on Cartesian grids can introduce computation over empty regions. Thus, PolarStream~\cite{chen2021} adopts polar coordinate representations. It uses polar pillars and multi-scale context padding from neighboring sectors for object detection and semantic segmentation. PARTNER~\cite{Nie_2023_ICCV} addresses polar feature distortion through attention-based realignment for 3D object detection, while PFCF~\cite{zhang2026mamba} combines a Mamba-based polar streaming backbone with a sector-feature buffer and a Cartesian full-scan pathway. Building on these LiDAR streaming foundations, we present, to the best of our knowledge, the first streaming semantic and panoptic segmentation framework for spinning FMCW radar.}\looseness=-1

%% file: sections/03_methodology.tex
\section{Technical Approach} \label{sec:technical-approach}

\begin{figure*}[h]
    \centering
    \includegraphics[width=0.95\textwidth]{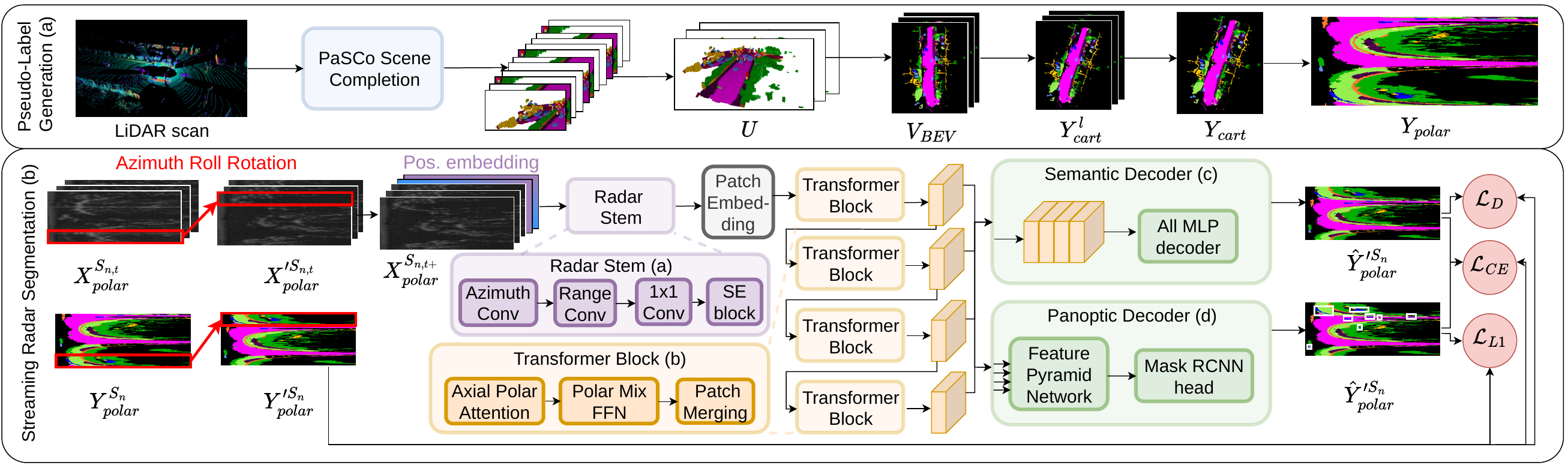} 
    \caption{\net~employs PaSCo scene completion to create dense radar labels, which are converted to a polar representation. We then apply azimuth roll rotation to augment radar scans before processing radar sectors with our streaming network. Key novelties include our RadarStem and Axial Polar Attention. Semantic and panoptic decoders decode the resulting features. Cross-entropy, Dice, and $L_1$ losses are applied per decoder, as appropriate.}
    \label{fig:method}
    \vspace{-0.3cm}
\end{figure*}

In this section, we present the methodology of \net~for generating dense semantic and panoptic radar annotations by leveraging a pre-trained LiDAR scene completion network. We also introduce a streaming-capable FMCW radar segmentation network that delivers semantic or panoptic predictions in real time, as shown in \figref{fig:method}.

\subsection{Cross-Modal Dense Label Generation}
We generate semantic and panoptic radar labels by leveraging cross-modal LiDAR supervision. Given a LiDAR point cloud $P \in \mathbb{R}^{K \times 4}$ of $K$ points $(x, y, z, i)$, we first produce a dense, scene-completed voxel grid using the LiDAR scene completion network PaSCo~\cite{cao2024pasco}.  
Specifically, PaSCo first applies a sparse 3D generative U-Net followed by a Mask2Former decoder~\cite{cheng2021mask2former} to produce panoptic scene completion (PSC) on occupied voxel regions.
We direct readers to the original work for implementation details~\cite{cao2024pasco}.
We employ PaSCo trained on SemanticKITTI~\cite{behley2019iccv}, which generates voxel grids $D \in \mathbb{R}^{256 \times 256 \times 32}$ at $0.2m$ resolution. 
To achieve extended spatial coverage while maintaining consistent voxel granularity, we decompose the LiDAR point cloud into quadrants and perform scene completion on each independently. Quadrant tiling is performed in voxel space to avoid overlap, and the resulting voxels are merged into a unified grid $U \in \mathbb{R}^{512 \times 512 \times 32}$. We then convert these 3D voxel representations to Cartesian BEV labels $V_{BEV}$ by collapsing along the z-axis and retaining, for each ground-plane cell, the label at the highest occupied voxel.

\textbf{Radar Projection and Alignment}: 
We project $V_{BEV}$ to radar coordinates via extrinsic calibration and nearest-neighbor interpolation to obtain a label map for each LiDAR scan in radar's Cartesian coordinate frame $Y_{cart}^{l}$.
A fundamental challenge in cross-modal supervision is temporal misalignment arising from differing sensor acquisition rates. Our FMCW radar operates at $\SI{4}{\hertz}$ while LiDAR operates at $\SI{10}{\hertz}$, creating temporal inconsistencies, particularly for dynamic objects. We address this through an azimuth-based synchronization strategy that mirrors the radar's rotating acquisition pattern. For each azimuth in a radar scan, we select the temporally nearest LiDAR scan $Y_{cart}^{l}$, effectively compositing $V_{BEV}$ from multiple consecutive LiDAR frames to generate $Y_{cart}$, a Cartesian label for every radar image.

Next, we convert the labels from Cartesian $Y_{cart}$ to polar coordinates $Y_{polar} \in \mathbb{R}^{A \times R}$ via range/azimuth resolution mapping, matching the native polar representation $X_{polar}$ of the radar image. Here, azimuth $A$ and range $R$ map to the image height and width, respectively.
We leverage polar representations for two reasons. First, prior radar segmentation work has shown that polar representations yield superior results compared to Cartesian bird's-eye-view (BEV) representations~\cite{chen2021}. Second, polar representations are particularly well suited to streaming: slicing along the azimuth (height) axis yields fully populated, rectangular slices. In contrast, slicing a Cartesian BEV scan produces sparse, wedge-shaped ``cake slices,'' where only a fraction of each slice contains valid data~\cite{chen2021}.

Although PSC provides instance-level panoptic annotations, we find them unreliable in the polar radar frame. When mapped to the sensor's coordinate system, instances are often oversized because small errors in the per-pixel estimates inflate the projected labels on the radar grid. Further, an object straddling the 0°/360° angular seam (e.g., spanning 350°-10°) has its pixels split across the two ends of the angular range. \looseness=-1

We derive panoptic labels by extracting instances from the polar semantic segmentation $Y_{polar}$ via connected-component analysis on same-class pixels. This can nonetheless produce fragmented or over-extended instances: a single object's return may split into several spatially disjoint components, or a projection inaccuracy may merge it into a thin streak spanning much of the frame. We address fragmentation and over-extension by restricting each instance to its single largest connected component, discarding disconnected fragments without altering any retained pixel. Further, we discard instances with a pixel area below $25$ or a bounding-box height below $4$, which typically correspond to spurious, non-physical detections.\looseness=-1

\subsection{Streaming Radar Segmentation}
This section describes streaming radar segmentation on our generated dataset $\mathbb{D} \in (X_{polar}, Y_{polar})$ (\figref{fig:method}b). We first slice $X_{polar}$ into sectors for streaming (\secref{sec:slicing}), followed by temporal frame aggregation to mitigate ghost objects (\secref{sec:temporal}). Then, we apply our novel azimuth roll augmentation which simulates ego-vehicle rotation during training (\secref{sec:rot}). Finally, we propose an encoder-decoder architecture with a radar-specific hierarchical Transformer encoder and task-specific decoders (\secref{sec:network}).\looseness=-1

\subsubsection{Polar Coordinate Slicing} \label{sec:slicing}
To leverage the continuous data streaming capability of FMCW radar, we process angular sectors sequentially during both training and inference, rather than accumulating complete $360^{\circ}$ scans. Since azimuth corresponds to the image height in our polar representation, we simply partition each polar input into $N$ sequential sectors, yielding $(X^{S_{n}}_{polar}, Y^{S_{n}}_{polar}) \in \mathbb{R}^{A/N \times R}$. 

\subsubsection{Temporal Frame Aggregation} \label{sec:temporal}
We fuse $t$ consecutive radar frames along the channel dimension, creating input $X^{S_{n,t}}_{polar} \in \mathbb{R}^{A/N \times R \times t}$. Since radar measurements suffer from multipath reflections, speckle noise, and sensor artifacts, temporal aggregation provides critical context to distinguish transient noise from structural information. We anchor $Y^{S_{n}}_{polar}$ to the last frame of the temporal sequence.

\subsubsection{Azimuth Roll Rotation Augmentation} \label{sec:rot}
Standard horizontal flipping is counterproductive for polar images due to range-dependent object representations. We instead introduce Azimuth Roll Rotation Augmentation, exploiting the periodic nature of the azimuthal axis to simulate ego-vehicle rotation via circular shifts along the image height. A random circular shift $s \sim \text{Uniform}[0, A)$ is applied along the azimuthal axis, with pixel wraparound at image boundaries. During training, we apply this augmentation with probability $p$.

\subsubsection{Network Architecture} \label{sec:network}
We design a tailored Mix Transformer radar-encoder for streaming polar inputs. Our encoder introduces a stem (RadarStem) that combines positional, directional, and temporal cues to capture range-dependent radar signatures. This is followed by anisotropic patch embeddings that retain angular resolution. Our novel axial polar attention then attends separately to the periodic azimuth axis and the bounded range axis, followed by a radar-adapted Mix-FFN. Finally, we append semantic and panoptic decoders. In contrast to the standard formulation of Mix Transformer~\cite{xie2021segformer}, which assumes Cartesian image geometry and doesn't model temporal continuity, our design captures the periodic azimuthal structure and streaming nature of polar radar sweeps.\looseness=-1

\textbf{RadarStem}: Polar images are not translation invariant because objects' representations depend on their range and azimuth from the ego-vehicle. We provide positional awareness by appending normalized embeddings $[-1, 1]$ to the $R$ and $A$ axes, producing $X^{S_{n,t+}}_{polar} \in \mathbb{R}^{A/N \times R \times (t+2)}$. This ensures the model understands that identical features at different ranges have distinct physical implications. 
Radar artifacts also exhibit directional characteristics such as azimuthal smearing and range-dependent reflection profiles. Standard dense convolutions are isotropic, making them unsuitable for capturing these anisotropic patterns. We employ asymmetric factorized depthwise convolutions with parallel radar-aligned filters where a $1 \times 5$ kernel captures range profiles and a $5 \times 1$ kernel captures azimuthal smearing. This approach reduces parameters and computation while maintaining receptive field coverage and improving feature expressiveness. 
We compress the $t+2$-channel representation to $t$ channels via $1 \times 1$ convolution and a Squeeze-and-Excitation (SE) block~\cite{hu2018squeeze}. The SE block applies channel-wise feature recalibration through global pooling and a bottleneck MLP. Since our channels encode temporal steps, this acts as a temporal attention mechanism, dynamically weighting the importance of historical versus current sweeps based on scene motion.\looseness=-1

\textbf{Patch Embeddings}: To compensate for the directional asymmetry of polar FMCW scans, we adopt anisotropic patch embeddings in the initial encoder stage~\cite{balvit25}. We substitute the standard $7 \times 7$ isotropic patch with a $3 \times 7$ rectangular patch (Azimuth $\times$ Range) and adjust the stride from $4 \times 4$ to $2 \times 4$ to match.\looseness=-1

\textbf{Axial Polar Attention}: Standard self-attention mechanisms for dense prediction rely on aggressive 2D spatial reduction for efficiency and assume flat Cartesian geometry. However, this corrupts fine-grained radar signatures and ignores its inherent periodicity.
We introduce Axial Polar Attention, a radar-tailored factorization that exploits the sensor's geometric characteristics. Azimuth (Height) Attention computes self-attention along the angular axis with circular padding on Key and Value tensors, explicitly modeling the 360° periodicity and enabling seamless attention across the 0°/360° boundary. In contrast, Range (Width) Attention applies standard zero padding to respect radar's fixed sensing bounds. This asymmetric padding strategy is novel to radar and fundamentally differs from standard axial attention.
Crucially, both branches are computed in parallel, and their outputs are summed to yield the final representation. This preserves computational efficiency for real-time streaming.

\textbf{Polar Mix-FFN}:
Positional information is implicitly encoded through convolutional zero-padding within the Mix-FFN layer of our encoder. Rather than the standard $3\times3$ convolution with uniform zero-padding, we apply asymmetric padding tailored to polar geometry: constant zero padding along the range axis to respect fixed sensing bounds, and circular padding along the azimuth axis to enforce angular periodicity. This wraps the image around the $0^\circ$/$360^\circ$ boundary, ensuring convolutional receptive fields accurately capture physical adjacency across the sweep boundary. This is only physically meaningful for the first and last sector under slicing, where the sector edge coincides with the true azimuth wraparound. For middle sectors, circular padding instead serves as a simplifying approximation, since these boundaries do not correspond to a genuine angular discontinuity.

\textbf{Semantic Decoder}: 
For decoding, we employ an All-MLP design that avoids expensive handcrafted components by relying solely on MLP layers. This is feasible given our Transformer encoder's large effective receptive field. The decoder unifies multi-level features, upsamples to $\frac{1}{4}$ of the input resolution, concatenates them, and applies MLPs to generate the segmentation mask.

For panoptic segmentation, the hierarchical Transformer generates feature maps at four spatial resolutions, $F_1$, $F_2$, $F_3$, and $F_4$, which we feed into a Feature Pyramid Network (FPN). For thing classes, we attach a Mask R-CNN head~\cite{maskrccn17}. The semantic branch upsamples each feature map until it reaches a uniform $\frac{1}{4}$ scale of the original resolution. The spatially aligned feature maps are summed element-wise, and a final linear projection predicts pixel-wise class probabilities across all class categories. Finally, we fuse the semantic and instance predictions using the same strategy as detailed in ~\cite{kirillov2019panoptic}.

\begin{table*}[t]
\centering
\footnotesize
\setlength\tabcolsep{3pt}
\caption{Streaming semantic segmentation performance on Boreas and SAX for full-scan (1) and streaming settings with varying numbers of sectors (2, 4, 8, 16, 32), in IoU [\%]. obj$_{L}$: large object categories (car, other-vehicle / obstacle). obj$_{S}$: small object categories (motorcycle, person, pole). bkg: background/static (road, pavement, terrain, vegetation, building, fence).}
\label{tab:streaming-seg}
\begin{tabular}{l|cccc|ccccc|cccc|ccccc}
\toprule
& \multicolumn{9}{c|}{Boreas} & \multicolumn{9}{c}{SAX} \\
\cmidrule{2-19}
Method & 1 & obj$_{L}$ & obj$_{S}$ & bkg & 2 & 4 & 8 & 16 & 32 & 1 & obj$_{L}$ & obj$_{S}$ & bkg & 2 & 4 & 8 & 16 & 32 \\
\midrule
TMVANet~\cite{Ouaknine_2021_ICCV} & 27.85 & 39.20 &2.47 & 45.69 & 25.87 & 23.06 & 24.13 & 24.77 & 19.66 & 32.27 & 36.76	& 6.55 & 47.71
& 24.37 & 22.34 & 20.98 & 20.78 & 17.53 \\
RSSNet~\cite{zhang24tarssnet} &  23.56	& 27.45	& 1.83	& 42.70 & 22.26	& 19.30 & 18.63 & 17.42 & 11.98 & 31.02	& 35.43	& 5.35 & 47.16 & 27.59 & 23.05	& 25.55	& 26.33	& \underline{20.14}\\
TransRadar~\cite{Dalbah_2024_WACV} &  30.13	&  43.34 & 2.60 & 48.85 & 27.74	& 28.17	& 26.18	& \underline{26.02}	& \underline{20.85} & 31.63	& 33.91	& 7.21 & 47.45& 32.67	& 28.75	& 28.00 & 26.15 & 18.88\\
Mask2Former~\cite{cheng2021mask2former} &  \underline{31.67} & \textbf{50.66} &  \underline{4.13}	& 46.54 & \underline{30.57} & \underline{30.03} & 26.42& - & - & \underline{33.56}	&  \underline{37.17}	& \textbf{8.33} &  \underline{49.97} & \underline{33.85}	& \underline{32.94}	& \underline{30.99} & - & - \\
SegFormer~\cite{xie2021segformer} &  29.53 & 39.60& 2.85 &  \underline{49.49} & 29.53 & 29.39 & \underline{28.55} & - & - & 30.88 & 36.81 & 5.76	& 46.29 & 29.83	& 30.24	& 30.17	& - & - \\
MobileNetv3~\cite{howard2017mobilenets} & 24.86	& 29.89	& 2.04	& 44.32 & 24.67	& 23.22	& 20.38	& 18.93	& 19.20 & 28.35	& 29.95	& 4.39	& 44.47 & 29.56	& 24.50 & 29.00 &	\underline{27.14}	& 16.48\\
\midrule
\rowcolor{Gray}\net(Ours) &  \textbf{31.73} & \underline{43.71} & \textbf{4.34}	& \textbf{51.12} & \textbf{31.85}	& \textbf{31.57}	& \textbf{30.09}	& \textbf{29.27}	& \textbf{21.82} & \textbf{34.35} & \textbf{41.94} &  \underline{7.64}	& \textbf{50.46} & \textbf{34.01} & \textbf{34.33}& \textbf{33.99}	& \textbf{32.95} & \textbf{23.23}\\
\bottomrule
\end{tabular}
\end{table*}

\textbf{Losses}:
For semantic segmentation, we optimize
\begin{equation}
\mathcal{L}_\text{sem} = \alpha\,\mathcal{L}_{CE} + \beta\,\mathcal{L}_{D},
\end{equation}
combining class-weighted Cross-Entropy $\mathcal{L}_{CE}$ and Dice loss $\mathcal{L}_{D}$.
For panoptic segmentation, we use
\begin{equation}
\mathcal{L}_\text{pan} = \mathcal{L}_{CE}^\text{sem} + \mathcal{L}_{CE}^\text{inst} + \mathcal{L}_{L1},
\end{equation}
where $\mathcal{L}_{CE}^\text{sem}$ is the semantic head's Cross-Entropy loss, $\mathcal{L}_{CE}^\text{inst}$ combines box classification and mask Cross-Entropy losses, and $\mathcal{L}_{L1}$ handles bounding box regression.

%% file: sections/04_experiments.tex
\section{Experimental Evaluation}
We evaluate \net~against nine baselines and provide a comprehensive ablation study to demonstrate the significance of our contributions. We first present an overview of the datasets used, followed by the experimental settings, performance results, and ablations.

\subsection{Datasets}
\label{sec:datasets}
We evaluate our approach on the SAX and Boreas datasets. 
\noindent{\textbf{SAX}: The SAX dataset \cite{SAX20} was collected in the UK using a Navtech CTS350-X FMCW scanning radar operating at \SI{4}{\hertz}, paired with a Velodyne HDL-64E LiDAR. The dataset comprises 20{,}875 radar scans with a maximum range of \SI{163}{\meter}, each rendered as a $400 \times 3768$ polar image. Since the LiDAR sensor matches SemanticKITTI, we directly apply PaSCo pretrained on SemanticKITTI within our auto-labeling approach. In addition to the generated labels, we hand-labeled $350$ LiDAR scans with \cite{segmentsai}, yielding $100$ annotated radar images spanning 12 classes: car, two-wheeled vehicle, other vehicle, person, rider, road, pavement, building, fence, vegetation, terrain, and pole.} \looseness=-1

\noindent{\textbf{Boreas}: The Boreas dataset \cite{burnett_ijrr23} was recorded in Canada with a Navtech CIR304-H scanning radar (\SI{4}{\hertz}) alongside a 128-channel Velodyne Alpha-Prime LiDAR. Since its LiDAR intensity distribution differs from that of the HDL-64E used in SemanticKITTI, we normalize Boreas intensities to $[0,1]$ prior to auto-labeling. The selected dataset contains 16{,}450 radar scans from four randomly chosen days with moderate-to-good weather conditions, each represented as a $400 \times 3371$ polar image and automatically annotated via auto-labeling into 8 classes: car, two-wheeled vehicle, person, obstacle, terrain, pole, pavement, and road. We merge the original lorry, truck, fence, and building classes into a single obstacle class because this sensor produces inconsistent LiDAR predictions.}

\begin{table*}[t]
\centering
\footnotesize
\caption{Streaming panoptic segmentation performance on Boreas and SAX datasets for full-scan (1) and streaming with 4 sectors. PQ: Panoptic Quality, PQ$_{\text{st}}$: PQ for stuff classes, PQ$_{\text{th}}$: PQ for thing classes, SQ: Segmentation Quality, RQ: Recognition Quality.}
\setlength\tabcolsep{4pt}
\begin{tabular}{l|ccccc|ccc|ccccc|ccc}
\toprule
& \multicolumn{8}{c|}{Boreas} &  \multicolumn{8}{c}{SAX} \\
\cmidrule{2-17}
Method
& \multicolumn{5}{c|}{1}
& \multicolumn{3}{c|}{4} 
& \multicolumn{5}{c|}{1}
& \multicolumn{3}{c}{4} \\
\cmidrule{2-17}
& PQ & PQ$_{\text{st}}$ & PQ$_{\text{th}}$ & SQ & RQ
& PQ & SQ & RQ
& PQ & PQ$_{\text{st}}$ & PQ$_{\text{th}}$ & SQ & RQ 
& PQ & SQ & RQ \\
\midrule
PanopticFPN~\cite{kirillov2019panoptic} & \underline{18.94}  & 22.57 & 12.88 &
\textbf{59.59} & \underline{27.94} & 20.23 & \underline{55.29} & 29.30 &  13.09 & \underline{19.22} & 4.51 & 55.47 & 19.48 &  13.44 & 57.44 & 19.67 \\
Mask2Former~\cite{cheng2021mask2former} & \underline{18.94} & \textbf{27.17} & 5.22 & 
47.35 & 27.63 & \underline{21.53} & 48.69 & \underline{30.57} & \textbf{15.45} & \textbf{22.58} & \underline{5.48} & 55.94 & \textbf{22.69} &  \underline{14.40} & \underline{62.06} & \underline{20.97} \\
OneFormer~\cite{jain23oneformer} & 17.35 & 24.03 & \underline{13.15} & 45.71 & 24.45 & 20.32 & 48.29 & 28.90 & 9.90 & 13.58 & 4.74 & \underline{56.01} & 14.38 & 11.68 & 52.90 & 16.85\\
\midrule
\rowcolor{Gray}\net{}(Ours) & \textbf{21.80} & \underline{25.47} & \textbf{14.47} & \underline{52.94} & \textbf{32.08} & \textbf{23.31} & \textbf{56.84} & \textbf{33.39} & \underline{13.37} & 18.88 & \textbf{5.66} & \textbf{56.86} & \underline{19.86} & \textbf{15.43} & \textbf{64.45} & \textbf{22.29}\\
\bottomrule
\end{tabular}
\vspace{-0.3cm}
\end{table*}

\subsection{Experimental Settings}
\label{sec:exp_settings}
We divide each dataset into training, validation, and test sets using a 70\%-15\%-15\% split, defined along distinct chronological and geographical regions to ensure an unbiased benchmark of generalization performance. We crop polar images to $400\times 1024$, discarding the range beyond this point, as our LiDAR-based labels extend only to $1024$. We use a temporal window of $t=3$ scans, together with roll rotation augmentation applied with probability $p=0.7$ and vertical flipping. Our Mix Transformer has feature dimensions $F_1$, $F_2$, $F_3$, and $F_4$ of $32$, $64$, $160$, and $256$. To address class imbalance, we set the loss weights $\alpha=1$ and $\beta=3$. We train for $80k$ iterations using AdamW with a batch size of $16$, weight decay of $0.01$, an initial learning rate of $10^{-3}$ for the backbone and $10^{-2}$ for the segmentation head. We apply a $1500$-iteration linear warm-up, followed by Multi-Step decay by a factor of $10$ at iterations $60k$ and $72k$.
To benchmark the effectiveness of our proposed method, we evaluate segmentation accuracy using mean Intersection over Union (mIoU), and panoptic performance using Panoptic Quality (PQ), Segmentation Quality (SQ), and Recognition Quality (RQ)~\cite{mohan25}. Additionally, we measure Frames Per Second (FPS) on an NVIDIA GeForce RTX 2080 Ti GPU.

\subsection{Quantitative Results}
In Tab.~\ref{tab:streaming-seg}, we compare \net{} with three radar-specific architectures and three general semantic segmentation models on full scans ($N=1$) and on sub-sectors of $2$, $4$, $8$, $16$, and $32$ sectors per scan, using each method's published code and settings. We modify radar-specific networks to use range-azimuth data only. Mask2Former relies on a frozen DINOv2 backbone, while the remaining models are trained end-to-end. We guide the Cross-Entropy loss with pre-computed class weights to address class imbalance, and tune the learning rate and batch size to maximize performance for each model. We do not evaluate Mask2Former and SegFormer at $16$ and $32$ sectors, as the resulting sector heights fall below the minimum spatial dimensions supported by their published implementations, and supporting them would require method-specific transformations altering the native sector representation.\looseness=-1

We observe that \net{} outperforms all baselines across all settings. While the margin at full-scan resolution is modest ($0.06$pp on Boreas and $0.79$pp on SAX), the benefit of our method is particularly evident at $8$ sectors, where we outperform the second-best method by $1.54$pp on Boreas and $3$pp on SAX. We attribute this to our axial polar attention and custom patch embedding, which remain effective even at smaller sector sizes. For SAX, performance is even maintained close to the full-scan level. We observe that all methods drop in performance as sector size decreases, since detecting objects becomes harder with insufficient context or only partial views. Nonetheless, \net{} remains robust, achieving above $21.8$ mIoU even when processing sectors as small as $12 \times 1024$ ($32$ sectors) which enables prediction rates of up to $128$\,Hz for a $4$\,Hz sensor. We further analyze the compute-efficiency of our method in comparison to the baselines in \secref{sec:ablation}.\looseness=-1

We further note that all methods achieve high scores on background classes such as road, pavement, terrain, vegetation, building, and fences on full scans (mIoU $>40$), while small, thin objects (obj$_S$) such as two-wheeled vehicles, persons, and poles consistently score below $10$ mIoU. We attribute this to their limited occurrence combined with the inherent difficulty of distinguishing small objects in radar, an imbalance that loss re-weighting alone could not resolve. Our axial attention's dedicated modeling of range and azimuth generally benefits small-object categories, except on SAX, where Mask2Former's per-instance queries and DINOv2 pretraining better preserve its especially fine-grained classes. Large objects (obj$_L$) perform reasonably well ($30$-$50$ mIoU). However, Mask2Former outperforms \net{} on large objects on Boreas, as its global, query-based cross-attention captures long-range coherence that our more localized, compute-efficient design achieves only to a limited extent.

Next, we compare our method on panoptic segmentation against PanopticFPN, Mask2Former, and OneFormer. Our method achieves the best PQ in three of four settings. On Boreas, we outperform by at least $1.78$pp PQ, driven primarily by improved RQ, i.e., better instance recognition, despite a slightly lower SQ at full-scan. We attribute this to our superior encoding of small objects via temporal consistency in our RadarStem and improved scene understanding through Axial Polar Attention. On SAX, our method outperforms on the four slicing setting but falls short of Mask2Former at full-scan, where it achieves equivalent thing-class PQ but a lower stuff-class PQ. We note that for most methods, PQ improves from one to four sectors, unlike semantic segmentation, which we attribute to improved performance on PQ$_{\text{st}}$. The improvement in PQ$_{\text{st}}$ under slicing is largely an artifact of the metric. Under slicing, the same region splits into smaller instances, each more likely to be scored as a correct match, yielding a systematically higher PQ$_{\text{st}}$ without a genuine gain in segmentation quality. \looseness=-1

\begin{figure*}
  \centering
  \footnotesize
  \setlength{\tabcolsep}{0.02cm}
  {\renewcommand{\arraystretch}{0.5}
    \begin{tabular}{p{0.17cm}p{2.83cm}p{2.83cm}p{2.83cm}p{0.17cm}p{2.83cm}p{2.83cm}p{2.83cm}}
    \multicolumn{4}{c}{Boreas} & \multicolumn{4}{c}{SAX} \\ \cmidrule{2-4} \cmidrule{6-8}
    & \multicolumn{1}{c}{Ground Truth} & \multicolumn{1}{c}{Baseline} & \multicolumn{1}{c}{\net~(ours)} &  & \multicolumn{1}{c}{Ground Truth} & \multicolumn{1}{c}{Baseline} & \multicolumn{1}{c}{\net~(ours)} \\
    a) & \includegraphics[width=\linewidth]{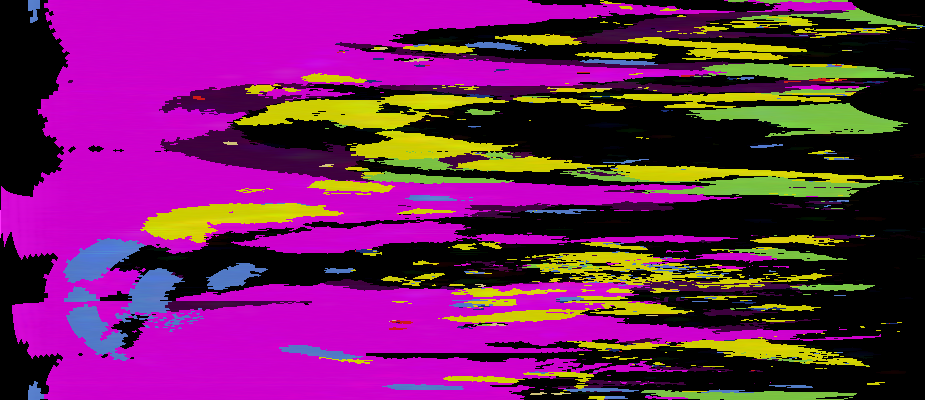} & 
      \includegraphics[width=\linewidth]{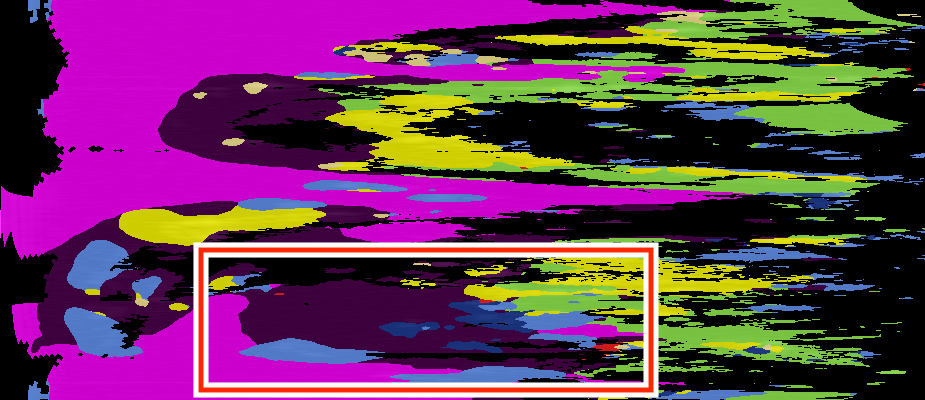} &
      \includegraphics[width=\linewidth]{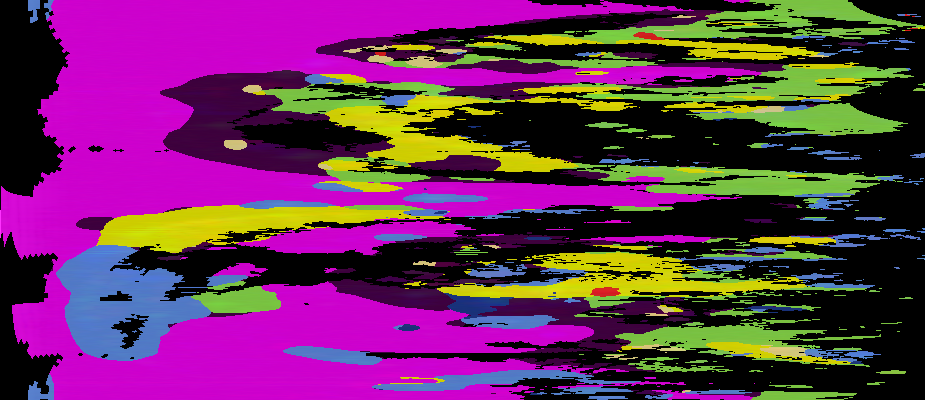} &
      e) &
      \includegraphics[width=\linewidth]{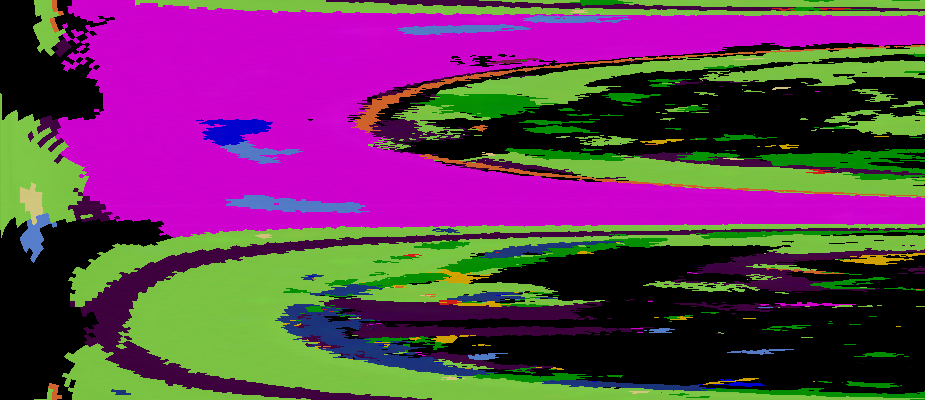} &  
      \includegraphics[width=\linewidth]{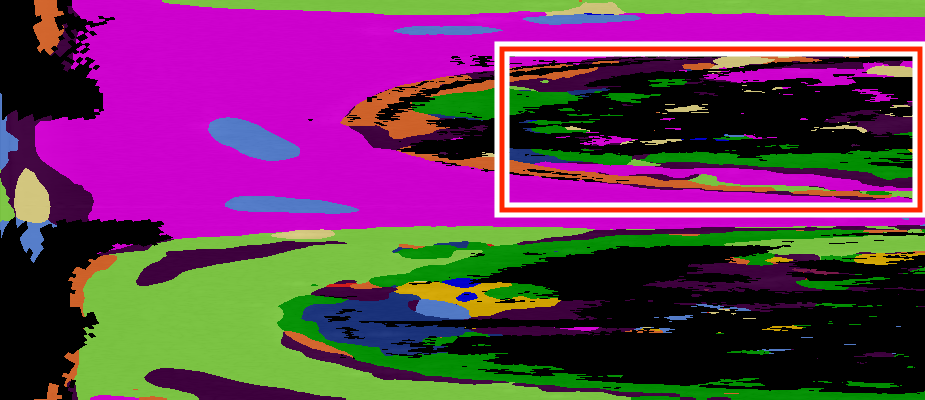} &
      \includegraphics[width=\linewidth]{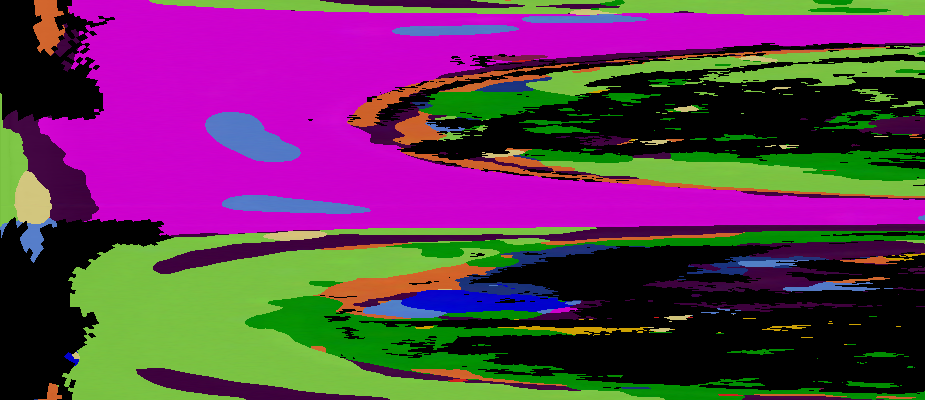} \\
    b) & \includegraphics[width=\linewidth]{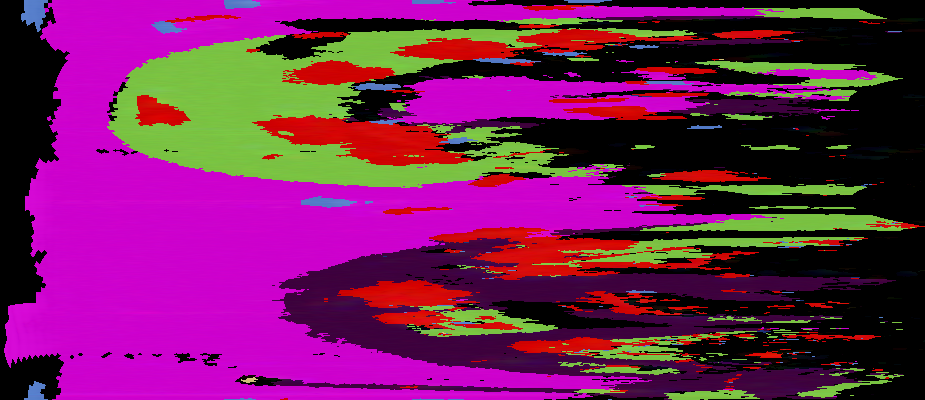} & 
      \includegraphics[width=\linewidth]{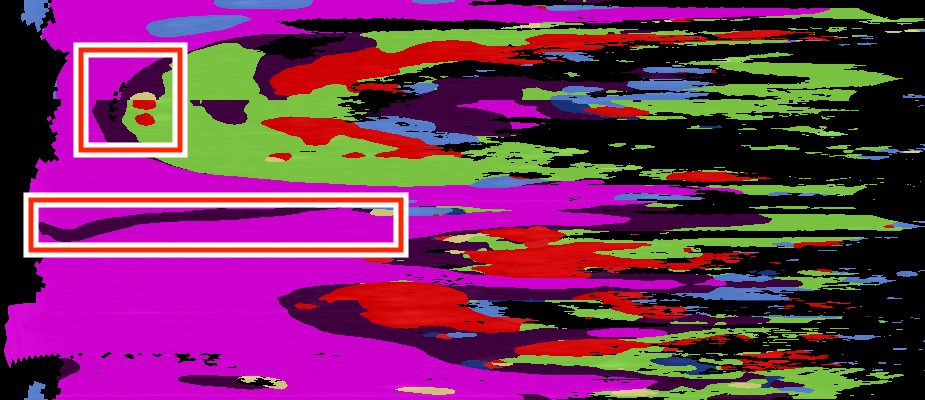} &
      \includegraphics[width=\linewidth]{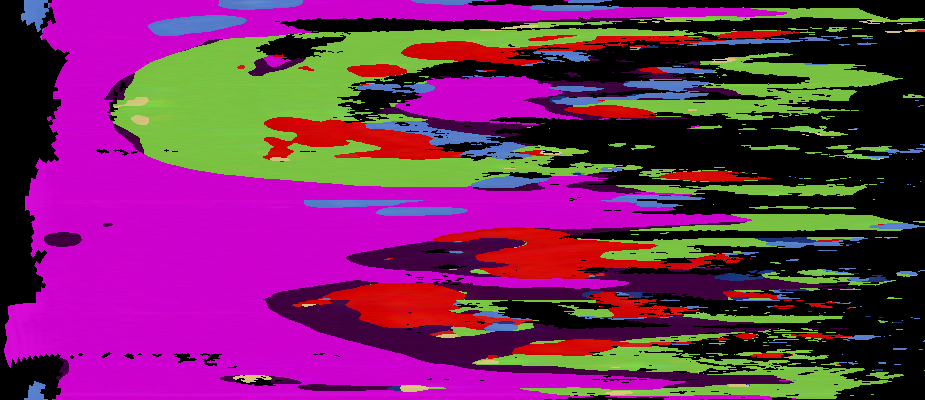} &
      f) &
      \includegraphics[width=\linewidth]{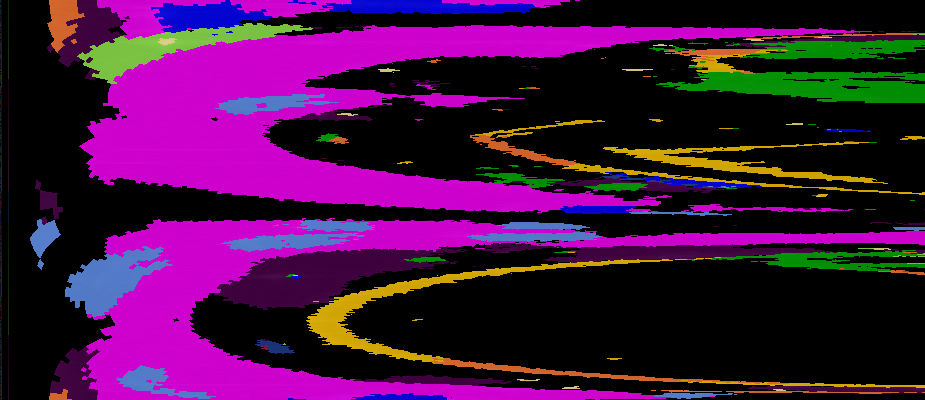} &  
      \includegraphics[width=\linewidth]{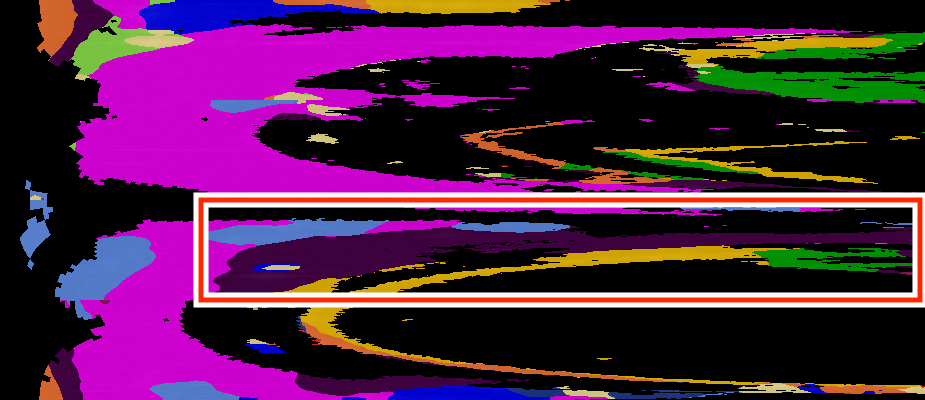} &
      \includegraphics[width=\linewidth]{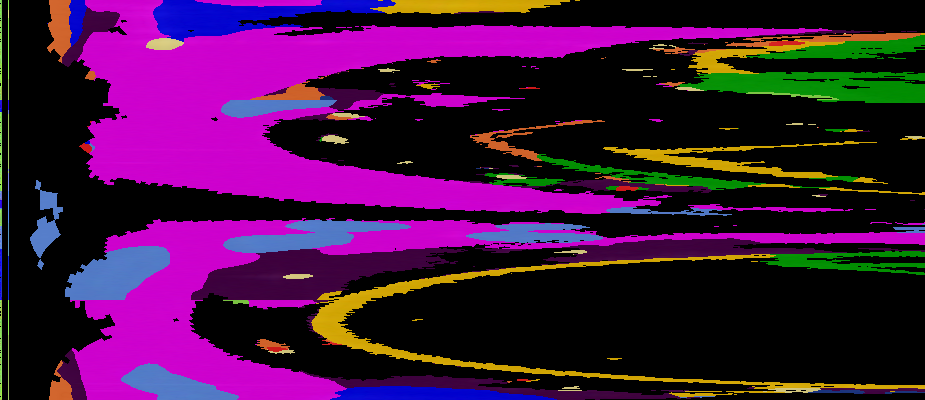} \\
    c) & 
      \includegraphics[width=\linewidth]{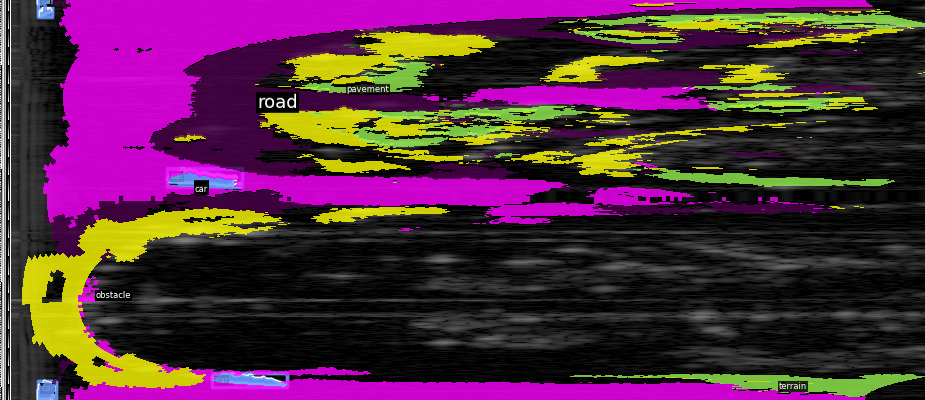} &
      \includegraphics[width=\linewidth]{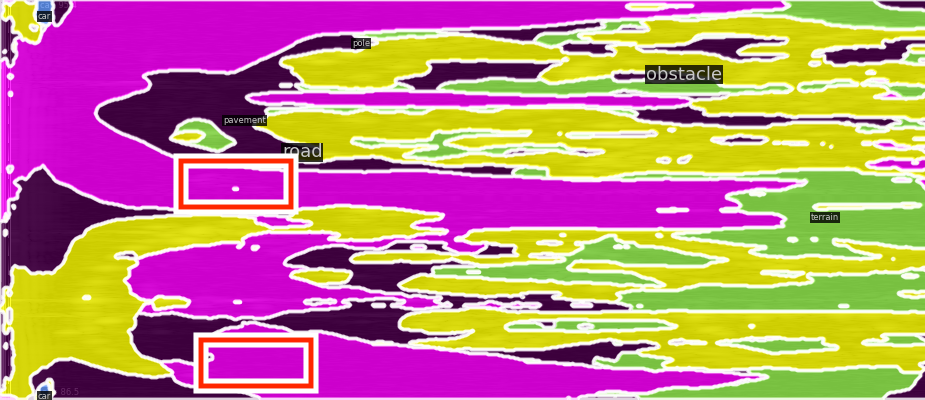} &
      \includegraphics[width=\linewidth]{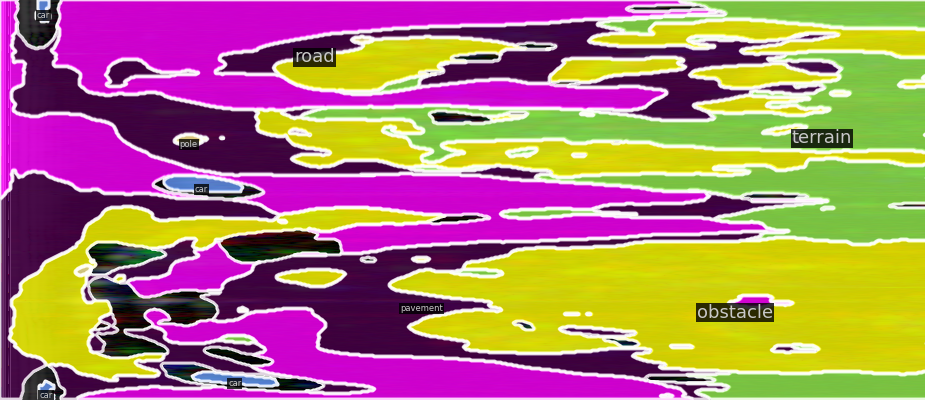} & g) &
       \includegraphics[width=\linewidth]{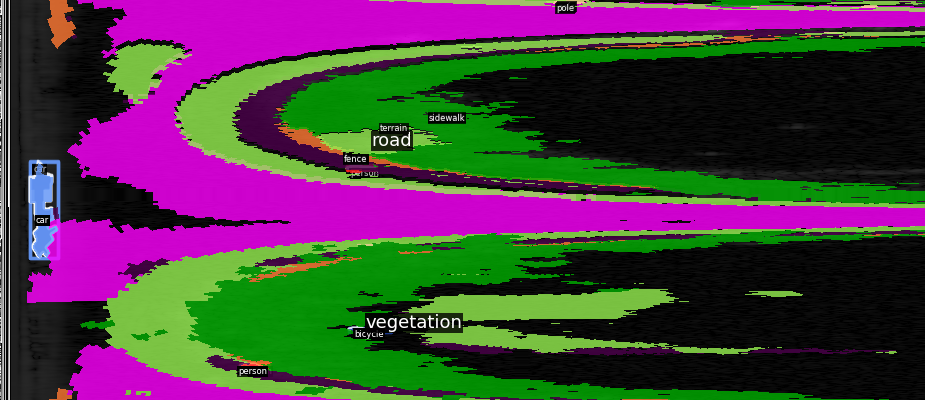} &
      \includegraphics[width=\linewidth]{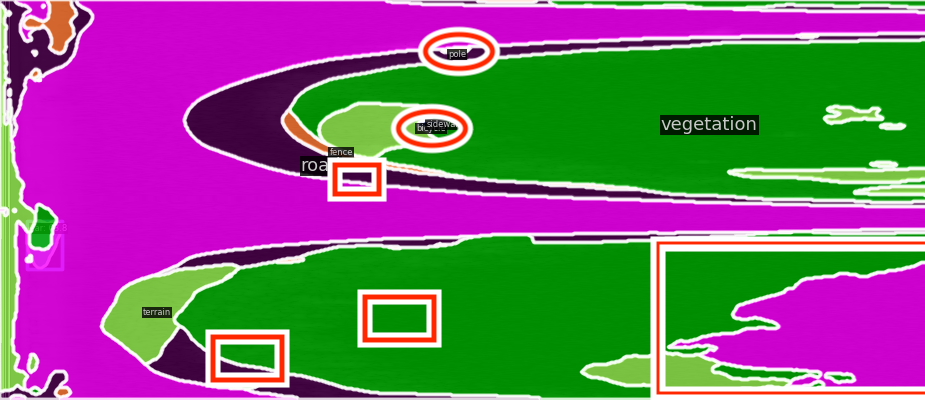} &
      \includegraphics[width=\linewidth]{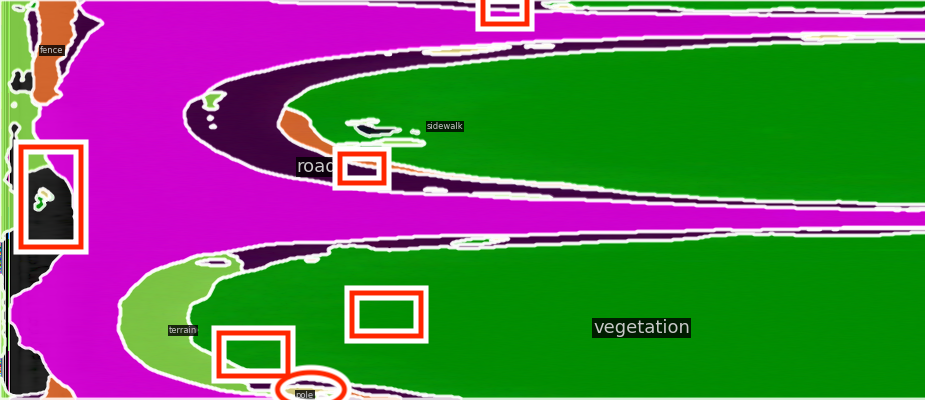} \\
    d) & 
            \includegraphics[width=\linewidth]{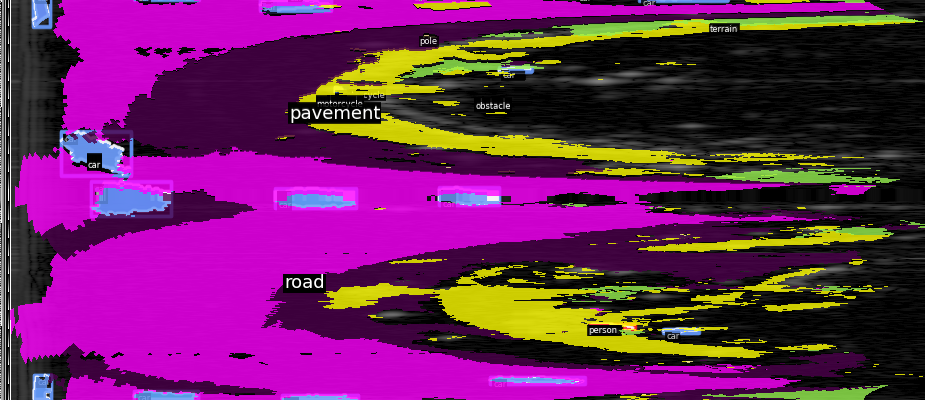} &
      \includegraphics[width=\linewidth]{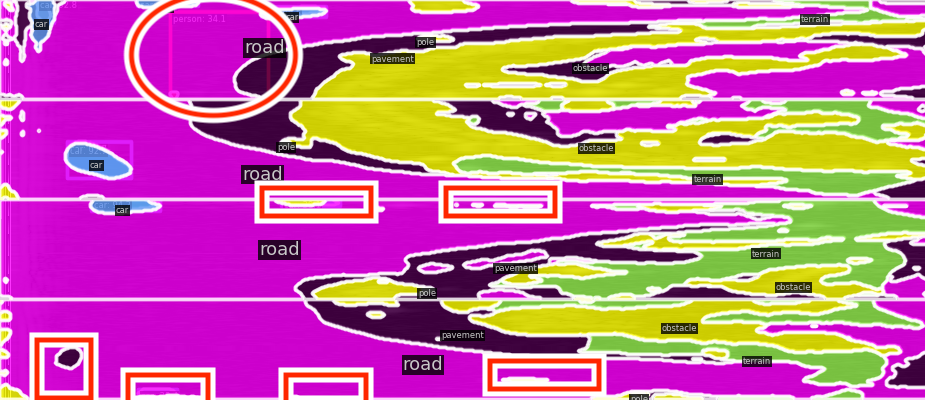} &
      \includegraphics[width=\linewidth]{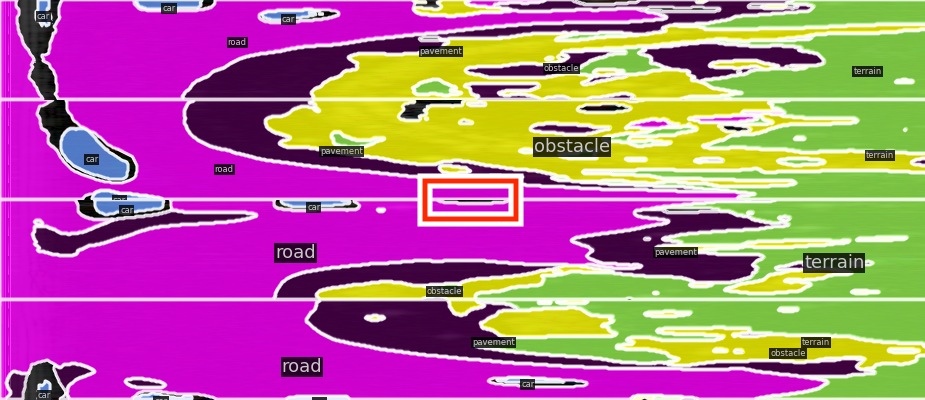} & h) &
       \includegraphics[width=\linewidth]{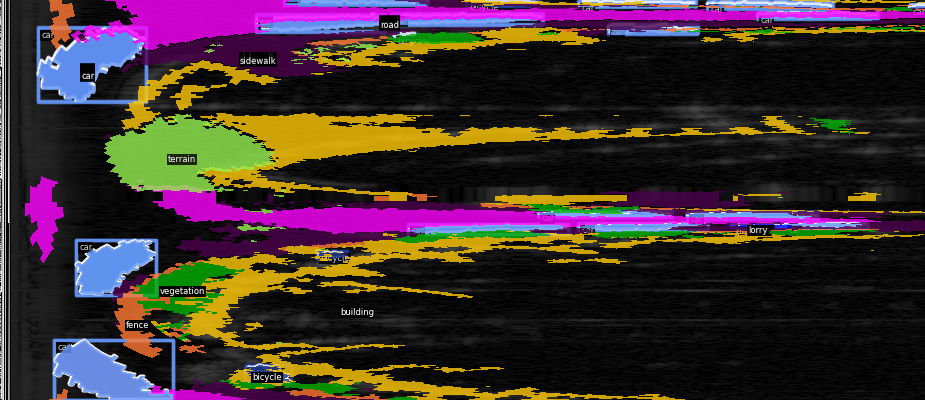} &
      \includegraphics[width=\linewidth]{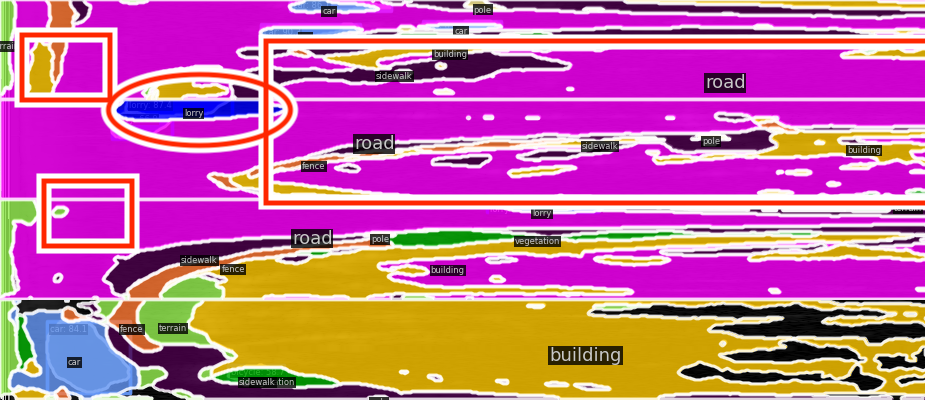} &
      \includegraphics[width=\linewidth]{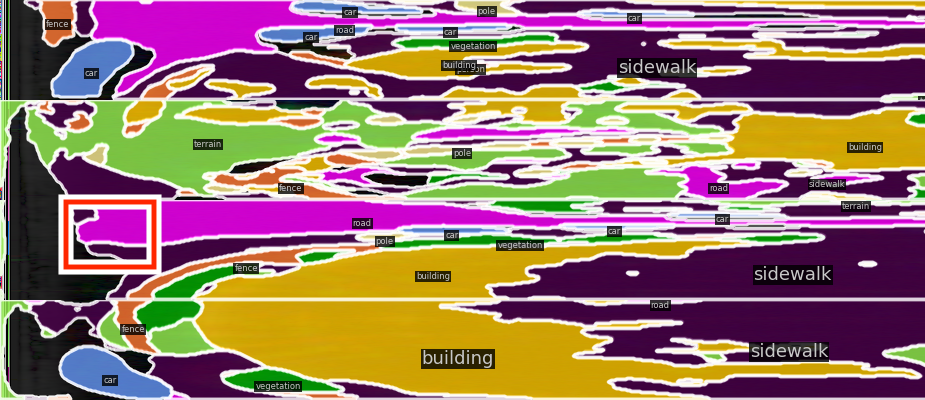} \\
    \end{tabular}}
  \caption{Qualitative results of \net~on SAX and Boreas. (a, e, c, g): full-scan (1 sector); (b, f, d, h): streaming (4 sectors). (a, b, e, f): Semantic Segmentation in comparison to SegFormer; (c, d, g, h): Panoptic Segmentation in comparison to Mask2Former. Errors are highlighted in white-red.}
  \label{fig:qualitative}
  \vspace{-0.3cm}
\end{figure*}

\subsection{Inference Performance}
Our reported inference speeds reflect network throughput measured with TensorRT and exclude sensor I/O and pre- and post-processing overhead. \tabref{tab:speed} reports inference speed (FPS), memory consumption, and latency percentiles. Mask2Former is significantly slower than the other tested methods in the full-scan and 4-sector streaming settings while SegFormer and \net~achieve comparable inference speeds at full-scan resolution. In the 4-sector setting, however, SegFormer attains the highest throughput as its spatial reduction attention scales down more favorably at small inputs than our dual-branch axial attention, whose fixed overhead from circular padding is amortized less effectively at small sector sizes. Despite this, \net~outperforms SegFormer by $4$pp while remaining highly efficient in absolute terms, comfortably exceeding real-time requirements even at the fine streaming granularity.
In terms of memory, Mask2Former requires over $9\times$ as much as our method, with its peak usage dominated by engine size and largely independent of input resolution in our setup, whereas RSSNet, SegFormer, and \net~all show reduced memory usage at smaller sector sizes. Across all methods, the tight spread between p50 and p99 latencies indicates stable, predictable inference times with minimal variation.\looseness=-1

\begin{table}[t]
\centering
\caption{Ablation study on the Inference performance on NVIDIA GeForce RTX 2080 Ti GPU, reporting throughput (FPS), memory footprint (Mem.[MiB]), and latency percentiles (p50-p95) on the SAX dataset full-scan and four sectors.}
\setlength\tabcolsep{3pt}
\label{tab:speed}
\begin{threeparttable}
\begin{tabular}{l|c|cccccc}
\toprule
Method & Sectors & IoU & FPS & Mem. & p50 & p90 & p99 \\
\midrule
\multirow{2}{*}{RSSNet \cite{prannay2020}} & 1 & 31.02 & 55.6 & 184 & 18.17 & 18.29 & 18.41 \\
 & 4 & 23.05 & 260.0 & 47 & 3.92 & 3.95 & 3.97 \\
\midrule
\multirow{2}{*}{Mask2Former \cite{cheng2021mask2former}} & 1 & 33.56 & 13.0 & 2080 & 77.15 & 78.02 & 78.30 \\
 & 4 & 32.94 & 46.4 & 2080 & 21.59 & 21.85 & 22.09 \\
\midrule
\multirow{2}{*}{SegFormer \cite{xie2021segformer}} & 1 & 30.88 & 145.8 & 299 & 7.06 & 7.10 & 7.14 \\
 & 4 & 30.24 & 782.3 & 38 & 1.34 & 1.35 & 1.36 \\
\midrule
\rowcolor{Gray} & 1 & 34.35 & 130.8 & 290 & 7.91 & 7.96 & 8.02 \\
\rowcolor{Gray}\multirow{-2}{*}{\net{} (Ours)} & 4 & 34.33 & 421.9 & 74 & 2.44 & 2.46 & 2.48 \\
\bottomrule
\end{tabular}%
\begin{tablenotes}[para,flushleft]
\footnotesize
\end{tablenotes}
\end{threeparttable}
\end{table}

\subsection{Ablation Study}\label{sec:ablation}
We ablate label generation, network components, and temporal window size on SAX for $1$ and $4$ sectors.

\subsubsection{Training Label Generation}
We show the influence of including PaSCo in our auto-labeling pipeline. \tabref{tab:training_data} compares training on auto-labels including PaSCo scene completion against sparse, pointwise labels predicted directly by its underlying backbone, WaffleIron~\cite{puy23waffleiron}. We evaluate both approaches on our hand-labeled test set and the dense auto-label test set. Our proposed auto-labeling approach substantially outperforms WaffleIron on the hand-labeled test set, improving results by at least $1.46$pp and showing that dense labels considerably improve prediction capabilities, while training with only a sparse signal hurts the network. Similarly, training on sparse WaffleIron labels generalizes poorly to the dense evaluation setting, with a gap of at least $9.2$pp, highlighting the importance of dense supervision for robust performance across label granularities.

\begin{table}
\centering
\caption{Performance comparison of different training data for \net{}.}
\label{tab:training_data}
\begin{threeparttable}
\begin{tabular}{l|cc|cc}
 \toprule
Training Label & \multicolumn{2}{c|}{Hand-Labeled Test } & \multicolumn{2}{c}{Dense Auto-Label Test} \\
Generation & 1 & 4 & 1 & 4 \\
\midrule
WaffleIron \cite{puy23waffleiron} & 29.07 & 28.76 & 25.06 & 23.59 \\
\rowcolor{Gray}PaSCo (Ours) & \textbf{31.47} & \textbf{30.22} & \textbf{34.35} & \textbf{34.33} \\ 
\hline
\end{tabular}
  \begin{tablenotes}[para,flushleft]
       \footnotesize
  \end{tablenotes}
\end{threeparttable}
\vspace{-0.3cm}
\end{table}

\subsubsection{Effect of Network Component}

The baseline SegFormer, processing a static single-channel polar radar image
(triplicated across RGB), achieves an mIoU of $30.88$ as shown in \tabref{tab:ab_elem}. Naively stacking
sequential frames \{$\tau$, $\tau-1$, $\tau-2$\} into a multi-channel input yields no
improvement, suggesting the network struggles to disentangle overlapping
spatial and temporal representations in sequential polar scans. Adding a
learnable stem module before the Transformer backbone, which concatenates
normalized range-azimuth grids $[-1, 1]$ as spatial priors, improves results
by \SI{0.77}{pp}. Azimuth roll rotation, a radar-realistic augmentation, adds
\SI{1.71}{pp}. Switching to rectangular patch embeddings contributes a
further \SI{0.71}{pp}. Finally, our radar-specific axial attention yields the
best overall result of \SI{34.35}{} mIoU, confirming the benefit of
tailoring attention to polar radar structure.

\begin{table}
\centering
\caption{Ablation study on various components of \net{} on the SAX dataset full-scan.}
\label{tab:ab_elem}
\begin{threeparttable}
\begin{tabular}{cccccc}
 \toprule
Temporal & Radar & Azimuth Roll & Patch & Axial Polar & \\
Frame & Stem & Aug. &Embed. & Att. & 1\\
\midrule
& & & & & 30.88 \\
\checkmark & & & & & 30.84 \\
\checkmark & \checkmark & & & & 31.61 \\
\checkmark & \checkmark & \checkmark &  &  & 33.32\\
\checkmark & \checkmark & \checkmark & \checkmark & & 34.03 \\
\rowcolor{Gray}\checkmark & \checkmark & \checkmark & \checkmark & \checkmark & \textbf{34.35} \\
\bottomrule
\end{tabular}
  \begin{tablenotes}[para,flushleft]
       \footnotesize
  \end{tablenotes}
\end{threeparttable}
\vspace{-0.3cm}
\end{table}

\subsubsection{Temporal Aggregation Window}
We evaluate different temporal windows in \tabref{tab:temporal}. Both a smaller ($t=1$) and larger temporal window ($t=5$) decrease performance by at least \SI{1.31}{pp}. Skipping intermediate frames decreases performance by at least $1.72$pp compared to using close windows, which we attribute to close frames combined with our novel axial attention helping detect ghost objects.

\begin{table}[t]
\centering
\caption{Ablation study on temporal frame aggregation of \net{} on SAX full-scan. $\tau$ denotes the timestamp of stacked radar frames.}
\label{tab:temporal}
\setlength\tabcolsep{4pt}
\begin{threeparttable}
\begin{tabular}{ccc|cc}
\toprule
 $\{\tau\}$ & $\{\tau,\tau\text{-}1,\tau\text{-}2\}$ & $\{\tau,\dots,\tau\text{-}4\}$ & $\{\tau,\tau\text{-}3,\tau\text{-}5\}$ & $\{\tau,\tau\text{-}4,\tau\text{-}7\}$ \\
\midrule
32.05 & \textbf{34.35} & 33.04 & 32.63 & 32.17 \\
\bottomrule
\end{tabular}
\end{threeparttable}
\vspace{-0.2cm}
\end{table}

\subsection{Qualitative Results}
We present qualitative results for semantic and panoptic segmentation in \figref{fig:qualitative}. For semantic segmentation, we compare with SegFormer, whose comparable FPS highlights the resulting quality disparity, and for panoptic segmentation, with the strongest baseline, Mask2Former. Panels (a, e) show full-image predictions on Boreas and SAX, respectively, while (b, f) show 4-sector streaming splits on the same datasets. In (a, b, f), SegFormer confuses street and sidewalk, whereas \net~predicts both classes correctly. Panel (b) further highlights the smooth transition between sectors achieved by \net{}, while in (e), our method correctly segments vegetation, whereas the baseline confuses it with sidewalk. 
For panoptic segmentation, we observe that Mask2Former commonly misses cars in (c, d) for Boreas while it commonly predicts street for sidewalk in (d) and building in (h). Further, both models struggle to detect small objects such as bicycle and person in (g). 
Additional qualitative results on Boreas under adverse weather conditions are provided in the multimedia material.\looseness=-1

%% file: sections/05_conclusion.tex
\section{Conclusion}\label{sec:conclusion}
We present \net{}, a novel approach to radar labeling and streaming segmentation. We first generate dense labels via LiDAR scene completion, yielding richer supervision for improved segmentation. We then introduce a lightweight, radar-specific transformer architecture for streaming semantic and panoptic segmentation, enabling fast sub-scan predictions. Our contributions include tailored augmentation, a radar-specific patch embedding and stem architecture, and a novel attention formulation that exploits range-azimuth polar representations. Experiments on the Boreas and SAX datasets show that our method outperforms others in both performance and compute requirements for panoptic and semantic segmentation across various sector sizes. This is the first work to address streaming and panoptic segmentation for FMCW radar. We hope our dataset, auto-labeling pipeline, and architecture will serve as a foundation for future research on real-time radar scene understanding.

%% file: references.bib
@String(CVPR = Proc # {IEEE Conf. Comput. Vis. Pattern Recog.})

@String(ICCV = Proc # {Int. Conf. Comput. Vis.})

@String(ECCV = Proc # {Eur. Conf. Comput. Vis.})

@String(NIPS = {Adv. Neural Inform. Process. Syst.})

@String(IROS = Proc # {IEEE Int. Conf. on Intel. Rob. and Syst.})

@String(ICRA = Proc # {IEEE Int. Conf. on Rob. and Auto.})

@String(RAL = {IEEE Rob. and Auto. Let.})

@inproceedings{sheeny2021radiate,
  title={Radiate: A radar dataset for automotive perception in bad weather},
  author={Sheeny, Marcel and De Pellegrin, Emanuele and Mukherjee, Saptarshi and Ahrabian, Alireza and Wang, Sen and Wallace, Andrew},
  booktitle=ICRA,
  pages={1--7},
  year={2021},
}

@inproceedings{zhang2026mamba,
  title={Polar Hierarchical Mamba: Towards Streaming LiDAR Object Detection with Point Clouds as Egocentric Sequences},
  author={Zhang, Mellon M and Chou, Glen and Mukhopadhyay, Saibal},
  booktitle={Proc. of the IEEE/CVF Winter Conf. on Applications of Computer Vision},
  year={2026}
}

@article{zeller2024semrafiner,
  title={SemRaFiner: Panoptic segmentation in sparse and noisy radar point clouds},
  author={Zeller, Matthias and Herraez, Daniel Casado and Ayan, Bengisu and Behley, Jens and Heidingsfeld, Michael and Stachniss, Cyrill},
  journal={IEEE Robotics and Automation Letters},
  volume={10},
  number={2},
  pages={923--930},
  year={2024},
  publisher={IEEE}
}

@article{li2025get,
  title={Get it for free: Radar segmentation without expert labels and its application in odometry and localization},
  author={Li, Siru and Hong, Ziyang and Chen, Yushuai and Hu, Liang and Qin, Jiahu},
  journal=RAL,
  volume={10},
  number={3},
  year={2025},
}

@INPROCEEDINGS{carrada20,
    author={Ouaknine, Arthur and Newson, Alasdair and Rebut, Julien and Tupin, Florence and Pérez, Patrick},
    booktitle={Int. Conf. on Pattern Recognition},
    title={CARRADA Dataset: Camera and Automotive Radar with Range- Angle- Doppler Annotations},
    year={2021},
    volume={},
    number={},
    pages={5068-5075},
}

@INPROCEEDINGS{prannay2020,
  author={Kaul, Prannay and de Martini, Daniele and Gadd, Matthew and Newman, Paul},
  booktitle={IEEE Intelligent Vehicles Symposium}, 
  title={RSS-Net: Weakly-Supervised Multi-Class Semantic Segmentation with FMCW Radar}, 
  year={2020},
  volume={},
  number={},
  pages={431-436}}

@InProceedings{frossard2021,
  title = 	 {StrObe: Streaming Object Detection from LiDAR Packets},
  author =       {Frossard, Davi and Suo, Shun Da and Casas, Sergio and Tu, James and Urtasun, Raquel},
  booktitle = 	 {Proc. of the Conf. on Robot Learning},
  pages = 	 {1174--1183},
  year = 	 {2021},
  volume = 	 {155},
}

@inproceedings{chen2021,
 author = {Chen, Qi and Vora, Sourabh and Beijbom, Oscar},
 booktitle=NIPS,
 pages = {26871--26883},
 title = {PolarStream: Streaming Object Detection and Segmentation with Polar Pillars},
 volume = {34},
 year = {2021}
}

@InProceedings{Nie_2023_ICCV,
    author    = {Nie, Ming and Xue, Yujing and Wang, Chunwei and Ye, Chaoqiang and Xu, Hang and Zhu, Xinge and Huang, Qingqiu and Mi, Michael Bi and Wang, Xinchao and Zhang, Li},
    title     = {PARTNER: Level up the Polar Representation for LiDAR 3D Object Detection},
    booktitle = ICCV,
    year      = {2023},
}

@article{burnett_ijrr23,
                        author = {Keenan Burnett and David J Yoon and Yuchen Wu and Andrew Z Li and Haowei Zhang and Shichen Lu and Jingxing Qian and Wei-Kang Tseng and Andrew Lambert and Keith YK Leung and Angela P Schoellig and Timothy D Barfoot},
                        title ={Boreas: A multi-season autonomous driving dataset},
                        journal = {The Int.Journal of Robotics Research},
                        volume = {42},
                        number = {1-2},
                        pages = {33-42},
                        year = {2023},
                        }

@InProceedings{ouaknine_2021_ICCV,
    author    = {Ouaknine, Arthur and Newson, Alasdair and P\'erez, Patrick and Tupin, Florence and Rebut, Julien},
    title     = {Multi-View Radar Semantic Segmentation},
    booktitle = ICCV,
    year      = {2021},
    pages     = {15671-15680}
}

@INPROCEEDINGS{heidbrink24,
  author={Heidbrink, Max and Sura, Oliver and Rangaraj, Vinoth Kumar and Reinecke, Marc and Hoffmann, Marcel and Vossiek, Martin},
  booktitle={European Radar Conference}, 
  title={Concept for Automatic Annotation of Automotive Radar Data Using AI-Segmented Camera and Lidar Reference Data}, 
  year={2024},
  volume={},
  number={},
  pages={292-295}}

@inproceedings{zhang24tarssnet,
title={{TARSS}-Net: Temporal-Aware Radar Semantic Segmentation Network},
author={Youcheng Zhang and Liwen Zhang and ZijunHu and Pengcheng Pi and Teng Li and Yuanpei Chen and Shi Peng and Zhe Ma},
booktitle=NIPS,
year={2024},
}

@ARTICLE{geo21ramp_cnn,
  author={Gao, Xiangyu and Xing, Guanbin and Roy, Sumit and Liu, Hui},
  journal={IEEE Sensors Journal}, 
  title={RAMP-CNN: A Novel Neural Network for Enhanced Automotive Radar Object Recognition}, 
  year={2021},
  volume={21},
  number={4},
  pages={5119-5132},
}

@InProceedings{Zhang_2023_CVPR,
    author    = {Zhang, Liwen and Zhang, Xinyan and Zhang, Youcheng and Guo, Yufei and Chen, Yuanpei and Huang, Xuhui and Ma, Zhe},
    title     = {PeakConv: Learning Peak Receptive Field for Radar Semantic Segmentation},
    booktitle = CVPR,
    year      = {2023},
}

@article{li2024adapkc,
  title={AdaPKC: PeakConv with Adaptive Peak Receptive Field for Radar Semantic Segmentation},
  author={Li, Teng and Zhang, Liwen and Zhang, Youcheng and Pi, Pengcheng and Lu, Zongqing and Liao, Qingmin and Ma, Zhe},
  journal=NIPS,
  volume={37},
  pages={136545--136575},
  year={2024}
}

@InProceedings{Dalbah_2024_WACV,
    author    = {Dalbah, Yahia and Lahoud, Jean and Cholakkal, Hisham},
    title     = {TransRadar: Adaptive-Directional Transformer for Real-Time Multi-View Radar Semantic Segmentation},
    booktitle = {Proc. of the IEEE/CVF Winter Conf. on Applications of Computer Vision},
    year      = {2024},
    pages     = {353-362}
}

@article{howard2017mobilenets,
  title={Mobilenets: Efficient convolutional neural networks for mobile vision applications},
  author={Howard, Andrew G and Zhu, Menglong and Chen, Bo and Kalenichenko, Dmitry and Wang, Weijun and Weyand, Tobias and Andreetto, Marco and Adam, Hartwig},
  journal={arXiv preprint arXiv:1704.04861},
  year={2017}
}

@article{xie2021segformer,
  title={SegFormer: Simple and efficient design for semantic segmentation with transformers},
  author={Xie, Enze and Wang, Wenhai and Yu, Zhiding and Anandkumar, Anima and Alvarez, Jose M and Luo, Ping},
  journal=NIPS,
  year={2021}
}

@inproceedings{kirillov2019panoptic,
  title={Panoptic feature pyramid networks},
  author={Kirillov, Alexander and Girshick, Ross and He, Kaiming and Doll{\'a}r, Piotr},
  booktitle=CVPR,
  year={2019}
}

@inproceedings{hu2018squeeze,
  title={Squeeze-and-excitation networks},
  author={Hu, Jie and Shen, Li and Sun, Gang},
  booktitle=CVPR,
  pages={7132--7141},
  year={2018}
}

@inproceedings{behley2019iccv,
  author = {J. Behley and M. Garbade and A. Milioto and J. Quenzel and S. Behnke and C. Stachniss and J. Gall},
  title = {{SemanticKITTI: A Dataset for Semantic Scene Understanding of LiDAR Sequences}},
  booktitle = ICCV,
  year = {2019}
}

@inproceedings{cao2024pasco,
  title={Pasco: Urban 3d panoptic scene completion with uncertainty awareness},
  author={Cao, Anh-Quan and Dai, Angela and De Charette, Raoul},
  booktitle=CVPR,
  pages={14554--14564},
  year={2024}
}

@misc{segmentsai,
 author = {Segments.ai},
 title = {Segments.ai data labeling platform},
 year = {2026},
 url = {https://segments.ai},
}

@ARTICLE{lisus25,
  author={Lisus, Daniil and Burnett, Keenan and Yoon, David J. and Poulton, Richard and Marshall, John and Barfoot, Timothy D.},
  journal=RAL, 
  title={Are Doppler Velocity Measurements Useful for Spinning Radar Odometry?}, 
  year={2025},
  volume={10},
  number={1},
  }

@InProceedings{han2020,
author="Han, Wei
and Zhang, Zhengdong
and Caine, Benjamin
and Yang, Brandon
and Sprunk, Christoph
and Alsharif, Ouais
and Ngiam, Jiquan
and Vasudevan, Vijay
and Shlens, Jonathon
and Chen, Zhifeng",
title="Streaming Object Detection for 3-D Point Clouds",
booktitle=ECCV,
year="2020",
pages="423--441",
}

@ARTICLE{venon22,
  author={Venon, Arthur and Dupuis, Yohan and Vasseur, Pascal and Merriaux, Pierre},
  journal={IEEE Trans. on Intelligent Vehicles}, 
  title={Millimeter Wave FMCW RADARs for Perception, Recognition and Localization in Automotive Applications: A Survey}, 
  year={2022},
  volume={7},
  number={3},
  pages={533-555},
}

@INPROCEEDINGS{fang2023,
  author={Fang, Shihong and Zhu, Haoran and Bisla, Devansh and Choromanska, Anna and Ravindran, Satish and Ren, Dongyin and Wu, Ryan},
  booktitle=ICRA, 
  title={ERASE-Net: Efficient Segmentation Networks for Automotive Radar Signals}, 
  year={2023},
  volume={},
  number={},
}

@InProceedings{Rebut_2022_CVPR,
               author = {Rebut, Julien and Ouaknine, Arthur and Malik, Waqas and P\'erez, Patrick},
               title = {Raw High-Definition Radar for Multi-Task Learning},
               booktitle = CVPR,
               month = {June},
               year = {2022},
               pages = {17021-17030}
               }

@INPROCEEDINGS{zou2023,
  author={Zou, Hao and Xie, Zhen and Ou, Jiarong and Gao, Yutao},
  booktitle=ICRA, 
  title={TransRSS: Transformer-based Radar Semantic Segmentation}, 
  year={2023},
  volume={},
  number={},
  pages={6965-6972}}

@INPROCEEDINGS{mohan25,
  author={Mohan, Rohit and Hindel, Julia and Drews, Florian and Gläser, Claudius and Cattaneo, Daniele and Valada, Abhinav},
  booktitle=IROS,
  title={Open-Set LiDAR Panoptic Segmentation Guided by Uncertainty-Aware Learning}, 
  year={2025},
  volume={},
  number={},
  pages={2224-2231}}

@INPROCEEDINGS{sax20,
  author={Gadd, Matthew and de Martini, Daniele and Marchegiani, Letizia and Newman, Paul and Kunze, Lars},
  booktitle={IEEE Intelligent Vehicles Symposium}, 
  title={Sense–Assess–eXplain (SAX): Building Trust in Autonomous Vehicles in Challenging Real-World Driving Scenarios}, 
  year={2020},
  volume={},
  number={},
  pages={150-155}}

@inproceedings{cheng2021mask2former,
  title={Masked-attention Mask Transformer for Universal Image Segmentation},
  author={Bowen Cheng and Ishan Misra and Alexander G. Schwing and Alexander Kirillov and Rohit Girdhar},
  booktitle=CVPR,
  year={2022}
}

@INPROCEEDINGS{balvit25,
  author={Hindel, Julia and Mohan, Rohit and Bratulić, Jelena and Cattaneo, Daniele and Brox, Thomas and Valada, Abhinav},
  booktitle=IROS, 
  title={Label-Efficient LiDAR Semantic Segmentation with 2D-3D Vision Transformer Adapters}, 
  year={2025},
  volume={},
  number={},
  pages={4115-4122}}

@INPROCEEDINGS{maskrccn17,
  author={He, Kaiming and Gkioxari, Georgia and Dollár, Piotr and Girshick, Ross},
  booktitle=ICCV, 
  title={Mask R-CNN}, 
  year={2017},
  volume={},
  number={},
  pages={2980-2988}}

@inproceedings{puy23waffleiron,
  title={Using a Waffle Iron for Automotive Point Cloud Semantic Segmentation},
  author={Puy, Gilles and Boulch, Alexandre and Marlet, Renaud},
  booktitle=ICCV,
  year={2023}
}

@inproceedings{jain23oneformer,
      title={{OneFormer: One Transformer to Rule Universal Image Segmentation}},
      author={Jitesh Jain and Jiachen Li and MangTik Chiu and Ali Hassani and Nikita Orlov and Humphrey Shi},
      booktitle=CVPR, 
      year={2023}
    }

@article{luz2026latent,
  title={Latent Gaussian Splatting for 4D Panoptic Occupancy Tracking},
  author={Luz, Maximilian and Mohan, Rohit and N{\"u}rnberg, Thomas and Miron, Yakov and Cattaneo, Daniele and Valada, Abhinav},
  journal={IEEE Robotics and Automation Letters},
  year={2026},
}
